# Asynchronous Distributed Genetic Algorithms with Javascript and JSON

Juan Julián Merelo-Guervós, Pedro A. Castillo, JLJ Laredo, A. Mora García, A. Prieto

*Abstract*— **In a connected world, spare CPU cycles are up for grabs, if you only make its obtention easy enough. In this paper we present a distributed evolutionary computation system that uses the computational capabilities of the ubiquituous web browser. Using Asynchronous Javascript and JSON (Javascript Object Notation, a serialization protocol) allows anybody with a web browser (that is, mostly everybody connected to the Internet) to participate in a genetic algorithm experiment with little effort, or none at all. Since, in this case, computing becomes a social activity and is inherently impredictable, in this paper we will explore the performance of this kind of virtual computer by solving simple problems such as the Royal Road function and analyzing how many machines and evaluations it yields. We will also examine possible performance bottlenecks and how to solve them, and, finally, issue some advice on how to set up this kind of experiments to maximize turnout and, thus, performance.**

## I. Introduction

Application–level networks (ALNs), are configured as a set of clients/servers (*servents*) that can provide their spare CPU cycles by means of a downloadable application, establishing a distributed computation network which can provide ad hoc computational power. Some ALN like SETI@Home have been quite successful [1], creating a virtual computer that has proccesed a good amount of teraflops, while other experiments such as Popular Power (and most others, in fact) have not [**?**].

The key feature of these application–level networks is the simplicity of use: we believe that the best way to obtain the participation of as many users as possible is to make it as very simple. In particular, it will be easier if they do not need to download a special application (such as a screen-saver) to participate, as is needed in BOINC, the sucessor to SETI@Home. For this reason, we are exploring the use of applications that are commonly installed in the user's computer, such as the web browser, which is available even in PDAs and some cellular phones[1]. Moreover, most browsers natively include a JavaScript interpreter [2], [3], [4] or virtual machine. JavaScript is an interpreted language[2], initially proposed by Netscape, and later adopted as an ECMA standard [5], [6], [7], [8]. In this way, most browsers are compatible, at least at a language level (not always at the level of browser objects, where there exists a reasonable compatibility, anyway). Most browser also include elements such as a Java virtual machine and a Flash plugin, which, with ActionScript, has more or less the same capabilities. However, there are several disadvantages to these: they might or might not be present (they are not *native*), they are *noisy* in the sense that, since they act as plugins, their execution is always noted by the user, their programs are more heavyweight than simple text code, and, finally, its integration with the browser is more awkward than the seamless integration that JavaScript offers. In any case, most things said here for JavaScript also apply to these and other plugins.

By itself, an interpreted languaje is not enough for creating a metacomputer if there is no way to convey information back from the client to the server in a seamless way. The ability to use the virtual machine included in browsers for distributed computing appeared with the XmlHttpRequest object, which allows asynchronous petitions to the server, in what has been called AJAX, Asynchronous JavaScript and XML [9]. AJAX is just one of the possible ways to perform asynchronous client-server communication, the others being AJAJ (Asynchronous Javascript and JSON), and *remoting* using applets or embedded objects. However, it is quite popular, and a wide user base and documentation is available for it, using any of these asynchronous client/server communication protocols. The traditional client/server model becomes then more egalitarian, or closer to a peer to peer model, since a bidirectional communication line appears: the browser can make calls to the server, do some computation and later send the results to the server.

AJAX (and AJAJ, which differ only in the way data is serialized) works as follows: the XmlHttpRequest is provided with a request to the server and a pointer to a *callback* function. The request generates an event, which is asynchronously activated when a reply is received making use of the *callback* function. Following this approach the browser is not locked, providing a way to program applications that are similar to the ones used at the desktop, in the sense that they do not have to wait for the application response to be loaded and rendered on the screen every time a request is made. It also means that a the user clicking on the Submit button is no longer needed to initiate communication with the server; any JavaScript thread can do so, with the constraint that the only server they can communicate with is the one that hosts the page the script is included in. On the other side, this provides a way to use the browser for application level networks that create distributed computing systems, since the request-response loop does not need the user participation in a fashion very similar to any other distributed computing application; these ALN can be

The authors are with the Departamento de Arquitectura y Tecnología de Computadores, University of Granada (Spain), corresponding email jj@merelo.net

[1]Whose computing power is similar to four-year-old desktop machines

[2]which has nothing to do with Java, other than the name and its syntactic similarity

controlled from the server with any programming language. Of course, it can also be combined with other distributed programming frameworks based on OpenGrid [10] or other distributed computing paradigms.

We previously used Ruby on Rails for making this kind of distributed AJAX application [11]; however, performance and scaling behavior were not too satisfactory mainly for two reasons: the nature of the Ruby on Rails server, which required the setup of a load-balancing server, and the (probably determinant) fact that part of the genetic algorithm was done on the server, with a low degree of paralellism and thus a high impact on performance. Latest experiments after publication yielded a maximum of 40 chromosome evaluation per second[3].

In this paper, following the same concept of distributed evolutionary computation on the browser via AJAX, we have redesigned the application using Perl and PostgreSQL (on the server) and Javascript, as before, on the client, and renamed it AGAJAJ (pronounce it A-gah-yai), which stands for *Asynchronous Genetic Algorithm with Javascript and JSON*; in this case, the genetic algorithm (solving the well known Royal Road problem) is carried out only on the clients, with the server used just for interchange of information among them. We will perform several experimentes in which clients donate computing power by just loading a web page to find out what kind of performance we can expect from this kind of setup, from the number of machines that will be made available by their users to the number of evaluations each one of them can perform; in these experiments, we have improved two orders of magnitude the performance achieved in the previous experiments which used Ruby on Rails, and also the number of concurrent machines available to perform them, showing that this kind of setup is ready to take more computing-intensive experiments without the need of an expensive server setup.

This paper follows our group's line of work on distributing evolutionary computation applications, which has already been adapted to several parallel and distributed computing paradigms (for example, Jini [12], JavaSpaces [13], Java with applets [14], service oriented architectures [15] and P2P systems [16], [17]). Evolutionary computation is quite adequate for this kind of distributed environment for several reasons: it is a population based method, so computation can be distributed among nodes (via distribution of population) in many different ways; besides, some works suggest that there are synergies among evolutionary algorithms and parallelization: isolated populations that are connected only eventually avoid the loss of diversity and produce better solutions in fewer time obtaining, in some cases, superlinear accelerations [18].

Of course, with a suitable work division method, many other algorithms could be adapted to browser-based distributed computation; however, in this paper will solve only genetic algorithms, and concentrate on raw performance, rather than algorithmic behavior.

[3]The published figure was even lower.

The rest of the paper is organized as follows: next section concentrates on the application of volunteer/involuntary computing to evolutionary computation; the setup is described in section III. Experiments and results are shown in section IV and discussed in V, along with future lines of work.

## II. State of the Art

So called *volunteer computing* [19], [20] systems are application-level networks set up so that different people can donate CPU cycles for a joint computing effort. The best known project is SETI@home[4], which, from the user's point of view, is a screen-saver which has to be downloaded and installed; when the user's CPU is not busy it performs several signal analysis operations. Some companies related to volunteer computing, such as Popular Power (and others; they are referenced, for example, in [21]) did some experimentation with Java based clients, but none has had commercial success; on the other hand, the SETI@Home program has been open-sourced and extended as the BOINC (Berkeley Open Infrastructure for Network Computing) framework [22]. This kind of volunteer computing has been adapted to evolutionary computation in several ocasions, using frameworks such as DREAM [23], which includes a Java-based virtual machine, GOLEM@Home, *Electric Sheep* [24] and G2-P2P [25]. Both approaches acknowledge that to achieve massive scalability, a peer to peer (P2P) approach is advisable, since it eliminates bottlenecks and single points of failure.

There are mainly two problems in this kind of volunteer networks: first of all, it is important not to abuse the CPU resources of volunteers; secondly, a sufficient number of users is needed in order to be able to do the required computation, which can be a problem on its own if there are too many of them, bringing the network, or at least the solution-collecting node, to its knees. A third problem is that performance prediction is difficult when neither the number of participants nor their individual node performances are known in advance. Finally, fault-tolerance [26] and cheating [27] are also important issues; if the environment is competitive, or any single computation is important, they will have to be taken into account.

In any case, we believe that the best way to obtain a good amount of users is to make it easy for them to participate, using technologies available in their computers, as the browser is. In fact, some suggestions were published (for example, the one of Jim Culbert in his weblog [28], and in some mailing lists), and, besides our own [11], there have been some recent papers and reports on similar setups. For instance, W. Langdon has been running for some time an interactive evolution experiment using Javascript in the browser [29], which was mainly intended for achieving high diversity in a fractal snowflake design than high performance. Even more recently, Klein and Spector [30] present a system based on the Push3 language, which is compiled to JavaScript

[4]See `http://setiathome.berkeley.edu/` for downloading the software and some reports.

in the browser. This system would be the closest to what we are presenting in this paper.

The proposed approach could also be considered as *parasitic computing* since, as stated in Section I, the only participation from the user will be to load a web page and click on a button; in fact, any AJAX-based could use these resources without his acquiescence (and, in any case, it would be desirable to run without causing much trouble). The concept was introduced by Barabási in [31], and followed by others (for instance, Kohring in [32]). In that work they proposed to use the Internet routers to compute a *checksum* by means of a set of specially crafted packets, whose aggregated result would be used to solve the SAT problem. Anyway, although the concept is interesting, there seems not to be a continuation for this work (at least openly), probably due to its inherent dangers (as analyzed in a paper by Lam et al. [33]).

The virtual machine embedded into the browser provides a way to easily do that kind of sneaky/parasitic computing, but JavaScript faces the handicap of being an interpreted language, which means that the efficiency of different implementations varies wildly. Moreover, it is not optimized for numerical computation but for object tree management (the so called DOM, document object model) and strings. Nevertheless its wide availability makes us think about considering it, at least as a possibility.

## III. Methodology and Experimental Setup

For this experiments we have designed and implemented a client-server program written in Perl (server-side) and Javascript (client-side), communicating using JSON via the XMLHttpRequest object. This object requires than the website and the AJAX requests are served from the same host, which is a constraint. Code for both is available, under the GPL, from `http://rubyforge.org/projects/dconrails/`. The algorithm runs on the client for a fixed number of generations, as shown in figure 1; running parameters are set from the server and are downloaded from it along with the webpage from which the experiment is run. A preset number of generations is run on the client, after which a request is made to the server with the best individual in the last generation. The algorithm stops and waits for the answer from the server. The server receives the request, stores it in a database, and sends back the best individual stored in the server. This individual is incorporated in the client population, which starts again to run. Several clients acting at the same time make requests asynchronously, using the facilities of the standard Apache web server. The server is thus used as a clearinghouse for interchange of information among the different clients; however, there's no explicit comunication or topology among the different nodes running the genetic algorithm. Besides, the fact that the server always contain the best individuals generated so far guarantees that the best solution (with a fixed number of evaluations resolution) available so far is always kept. The server also sends back the number of generations the client should run; which is usually the same number as before, but turns to 0, thus stopping the client, when the stopping condition is met.

Clients leave the experiment by the expeditive method of surfing away to another page or closing the web browser; in tabbed browsers (most browsers nowadays), a tab (or several) can run the experiment while the browser is available for other tasks. When the experiment has been running for a predeterminad number of evaluations (which were set, for this experiment, to 750000), all clients get a message to stop running, and change their user interface to a message offering them to reload the (new) experiment and start all over again. Besides, there is a *watching* daemon running on the server which checks the database for the number of individuals evaluated, and resets the experiment by incrementing the experiment ID by one and eliminating the population. Thus, experiments can run unchecked on a server while this watchgad daemons is running. Several additional utilities are also provided via several webpages, that inform on the state of the experiment, or allow to set the GA parameters. Experimental subjects were gathered by several methods: sending it via email to department and project coworkers, using the URL for the experiment as a Google Talk status line, as a Twitter (`http://twitter.com`) message, as a blog post, and, eventually, it was picked up by a wildly popular Spanish blog [5], which managed to gather the highest number of machines.

The experiment consisted in optimizing the 256-bits Royal Road function, and each instance consisted in a maximum of 750000 evaluations (which were barely enough to find the solution). The algorithm was steady state (with incorporation of the inmigrant every 20 generations), with rank-based selection and substitution; every generation, 50% of the population was generated, substituting the worst 50% individuals. Crossover priority was set to 80%, and mutation to 20%, changing 1% of the bits. However, these settings will have no influence on performance, other than the fact that, if the solution is found before the end of the experiment, the users will get bored and change to a new page[6].

Data was gathered from two different sources: the watcher-daemon logs, which mainly gave data about the number of individuals evaluated and the time needed for each experiment, and the Apache daemon log; the relevant lines were extracted just by using grep. It should be noted that the server was not running exclusively the experiment, but doing it along with the usual tasks. The server was a 700-MHz, 1 Gigabyte-RAM machine, with the database in another dual processor, 450-MHz machine. Both machines were running obsolete RedHat 7.x and 9.x Linux operating systems[7].

Results of the set experiments will be commented in the next section.

[5] Who posted it at `http://www.microsiervos.com/archivo/ordenadores/experimento-computacion-distribuida.html`.

[6] And this is just an example of how social factors in this kind of experiments affect performance.

[7] Both machines host our group web server and home pages; we thought it was better to run the experiment in our standard setup instead of a dedicated one.

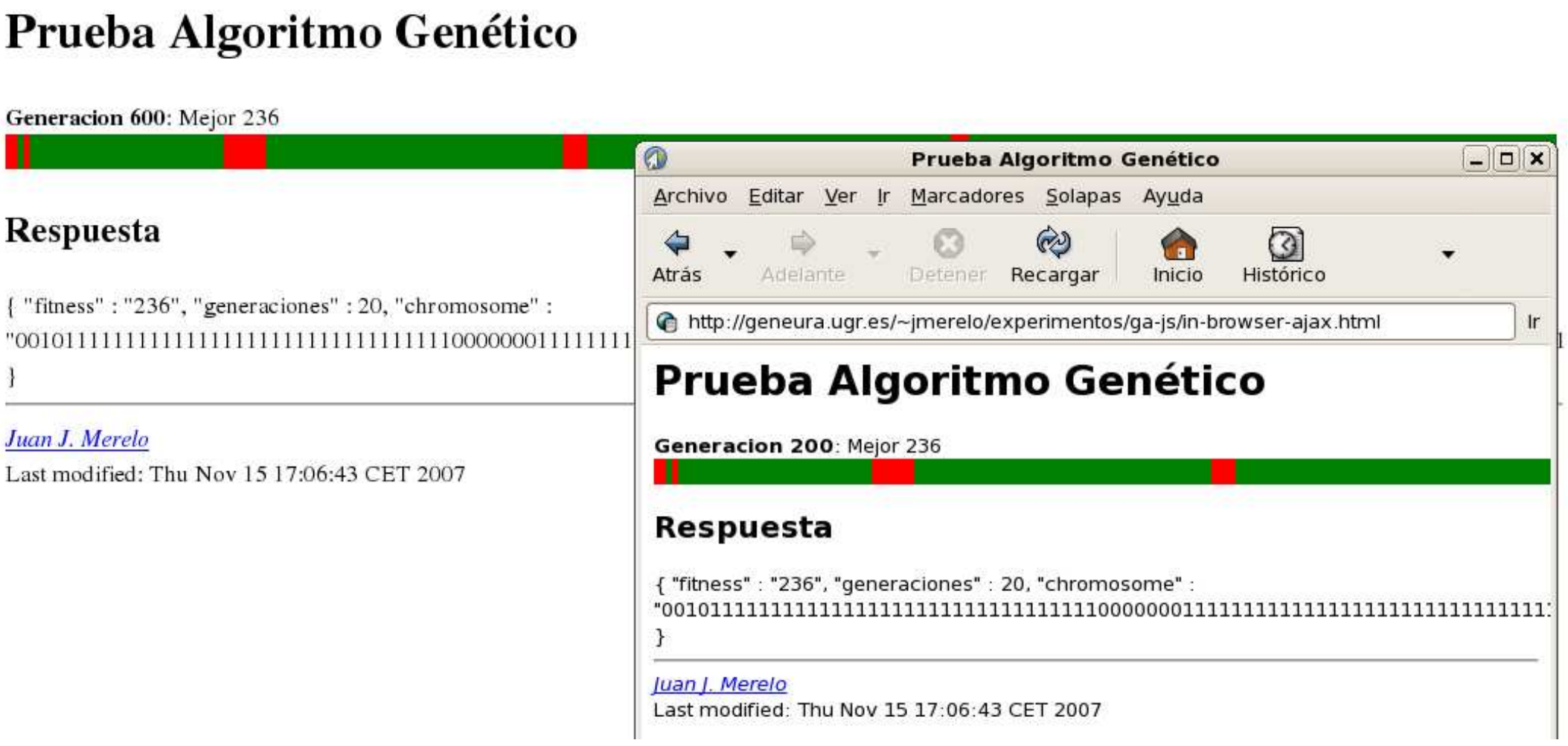

Fig. 1. Experiment running on two different browsers (Firefox and Epiphany) in the same machine. User interface is written in Spanish, since in this initial experiment was addressed to audience speaking that language. The colored (or gray-scale) horizontal bar is a graphical representation of the chromosome. The inset windows (Epiphany) started in second place, and thus, the state of evolution is less advanced.

## IV. Experimental results

Eventually, the experiment was running for several days, with different degrees of intensity. Several hundred machines participated in different instances, coming from all over the world, although mainly from Spain. The first question we wanted to answer was, how many machines should we expect in this kind of experiment? BOINC and SETI@home have gathered thousands of machines in its 10+ year run, but our experiments were limited in time (several hours, at most, if no machine was available, a few minutes if it was), so a limited number of machines should also be expected. The distribution of the number of machines is shown in figure 2.

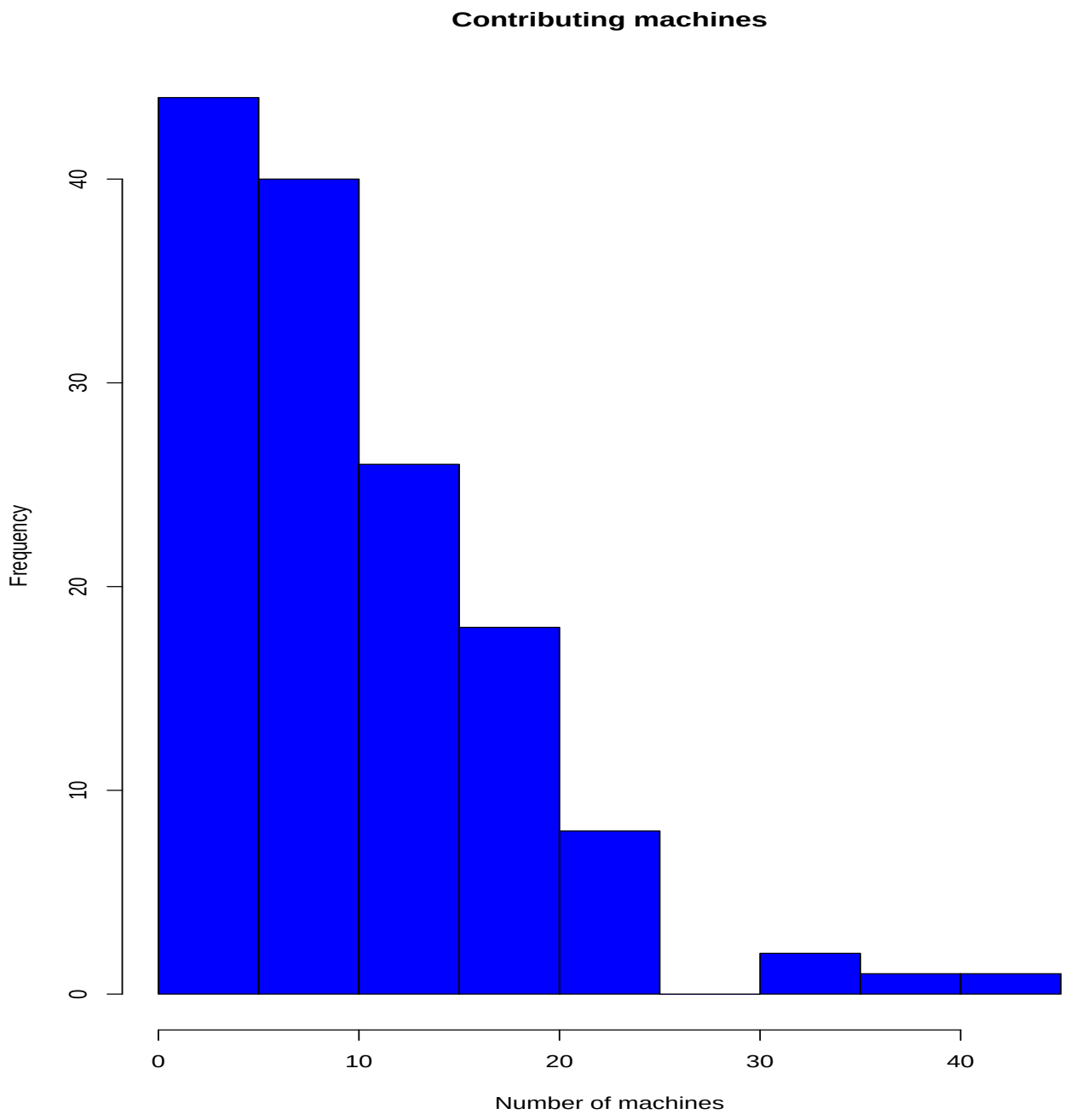

Fig. 2. Histogram of the number of machines used in each experiment. The median is 8.5 machines, and the 75% quartile is placed at 15 machines, with a peak of 44 machines. A single client using several browsers counts as a single machine. It should be noted that not all clients are simultaneously connected.

The exact figures will vary in every experiment, but it seems clear that the median number of machines will hover around 1/4 of the maximum number. Besides, it is quite easy to obtain 5 clients for a certain number of evaluations; most experiments have less than 10 clients.

On the other hand, the number of evaluations each client contributes are quite different, as is shown in figure 3.

Most clients contribute just a few generations, be it because the browser stops running the program if it takes more than a predetermined number of seconds (which can happen in IE or in Firefox if clients are too slow; usually the predetermined number of generations will be less than this number), the user gets bored and moves on, or because statistically most people join when the experiment has been running for some time and show up only when there are a few evaluations left. Each one of these problems would have to be tackled separately, but the baseline is that, even when a good number of machines joins an experiment, they will do so only for a limited amount of time on average. Besides, these limitations in number of generations translate also to limitations in time, so that experiments will have to be designed in a way that sufficient information is transmitted from clients during the expected time they are going to spend in the experiment.

But different clients have different performance, and it is also interesting to measure what is the average time (and thus average performance) it takes the clients between two communications to the server (that is, 20 * 50 evaluations +

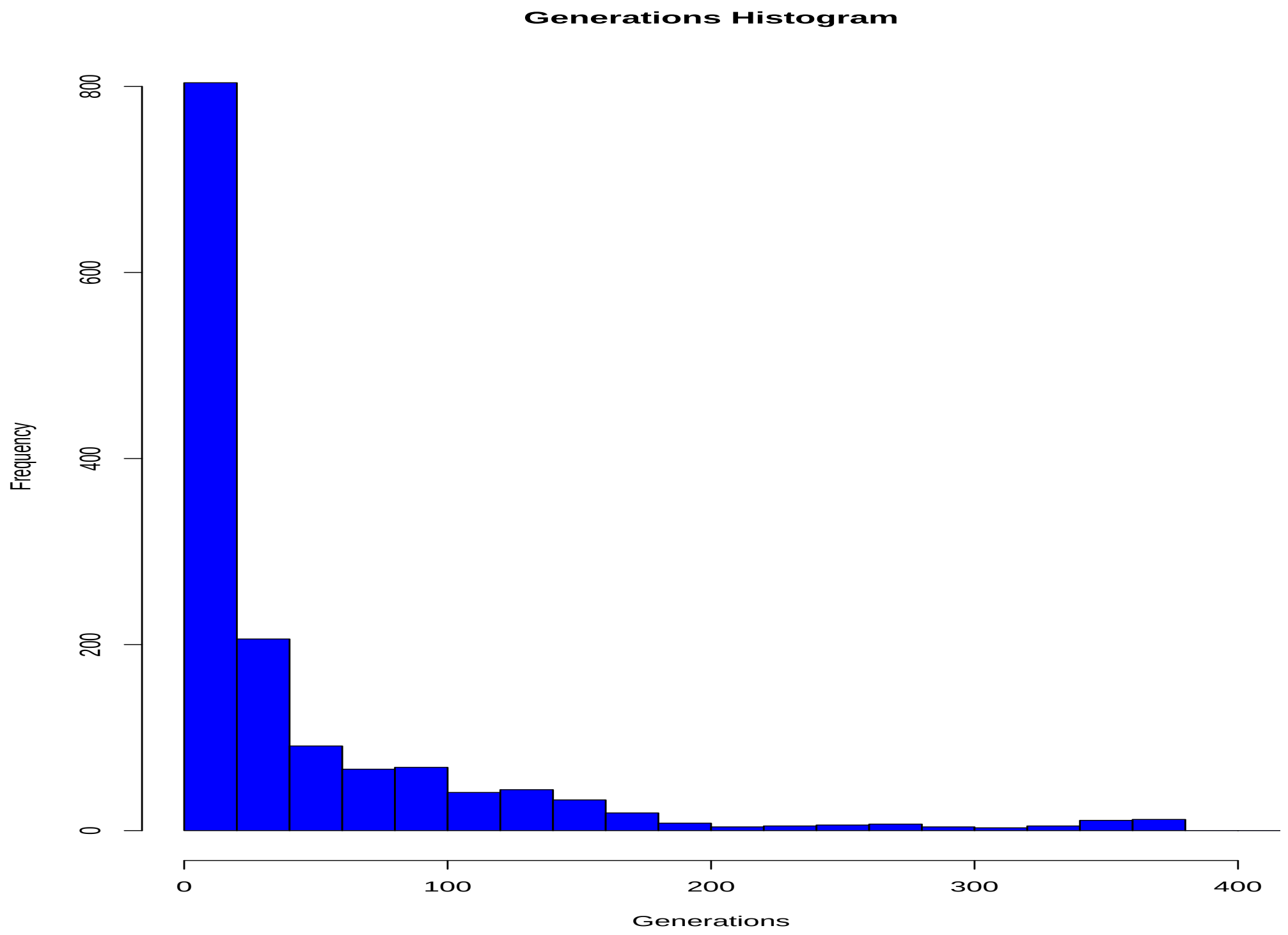

Fig. 3. Histogram of the number of generations all machines participating in the experiment have contributed, cut off at 400 generations. Each generation corresponds to 50 new individuals. Most clients contribute 100 generations or less, with a few contributing more than 200. The median is 16 generations, with the 3rd quartile placed at 55 generations.

waiting time). This is interesting for two main reasons: server performance will have to be tuned to be able to answer to this level of requests, and second, the generation gap will also have to be fine-tuned so that waiting time and the possibility that the script is blocked due to overtime is minimized. The results obtained in the experiment are shown in figure 4.

This figure shows that, for this kind of problem, the vast majority of clients will have a gap smaller than two seconds. This quantity will vary for different problems, but the conclusion is that most clients will have high or median performance, with few clients having lower performance. This measure also gives us an estimate of the average performance (2.906 seconds/20 generations).

However, at the end of the day the setup is intended to achieve high performance when running an evolutionary computation experiment. This data is presented in figure 5.

This figure is quite similar to fig 4. Median is at 1000 seconds, with a minimum at 292 and 3rd quartile at 2323; 75% of runs will last less than 2323 seconds. Taking into account that the average 20-generation step is 2.906 seconds, and a single experiment needs 375 such steps, the average single machine run would take 1089.75 seconds; this means that the maximum speedup achieved is 1089.75/292 = 3.73 clients running simultaneously, and the median is approximately a single *average* machine. This will probably vary for experiments of different duration, but, on average, we could say that significative (albeit small) speedups can be achieved using spontaneous volunteer computing. In general, however, several machines will sequentially provide CPU cycles to an experiment, adding up to a single machine doing all the work. In general also, the fact that there are up to 44 machines working in a single experiment, or that the range of running times can vary in a factor of up to one hundred, indicates that, for this experiment, no bottleneck has been found. Of course, more simultaneous machines will have to be tested to find the limit. Finally, the fact that all contributions are volunteer means that the evaluation rate is not constant, yielding figures like fig 6, where the steepness of each line is roughly equivalent to the evaluation speed, since the $x$ axis corresponds to time, and the $y$ axis number of individuals evaluated.

## V. Conclusions, discussion and future work

While in previous papers [11] we proved that this kind of AJAX based, volunteer, and potentially sneaky, computation could be used profitably for performing genetic algorithm experiments, in this paper we have proved that, without an expensive or far-fetched setup, it can achieve high performance, equivalent, at most, to several computers of average performance. The code used to perform the experiment is publicly available and is modular so that creating different experiments is just a matter of writing a new JavaScript fitness function and tuning the GA parameters accordingly.

The experiments have proved that there is a good amount of computational power that can be easily tapped and used for evolutionary computation experiments, however, the nature of AGAJAJ constrains also the way users donate computing power, as well as the number of clients available for an experiment. In this paper we have found some figures, which will undoubtedly vary for other experiments; however, the general shape of the curves will probably be the same, following a very steep decrease from the maximum values obtained.

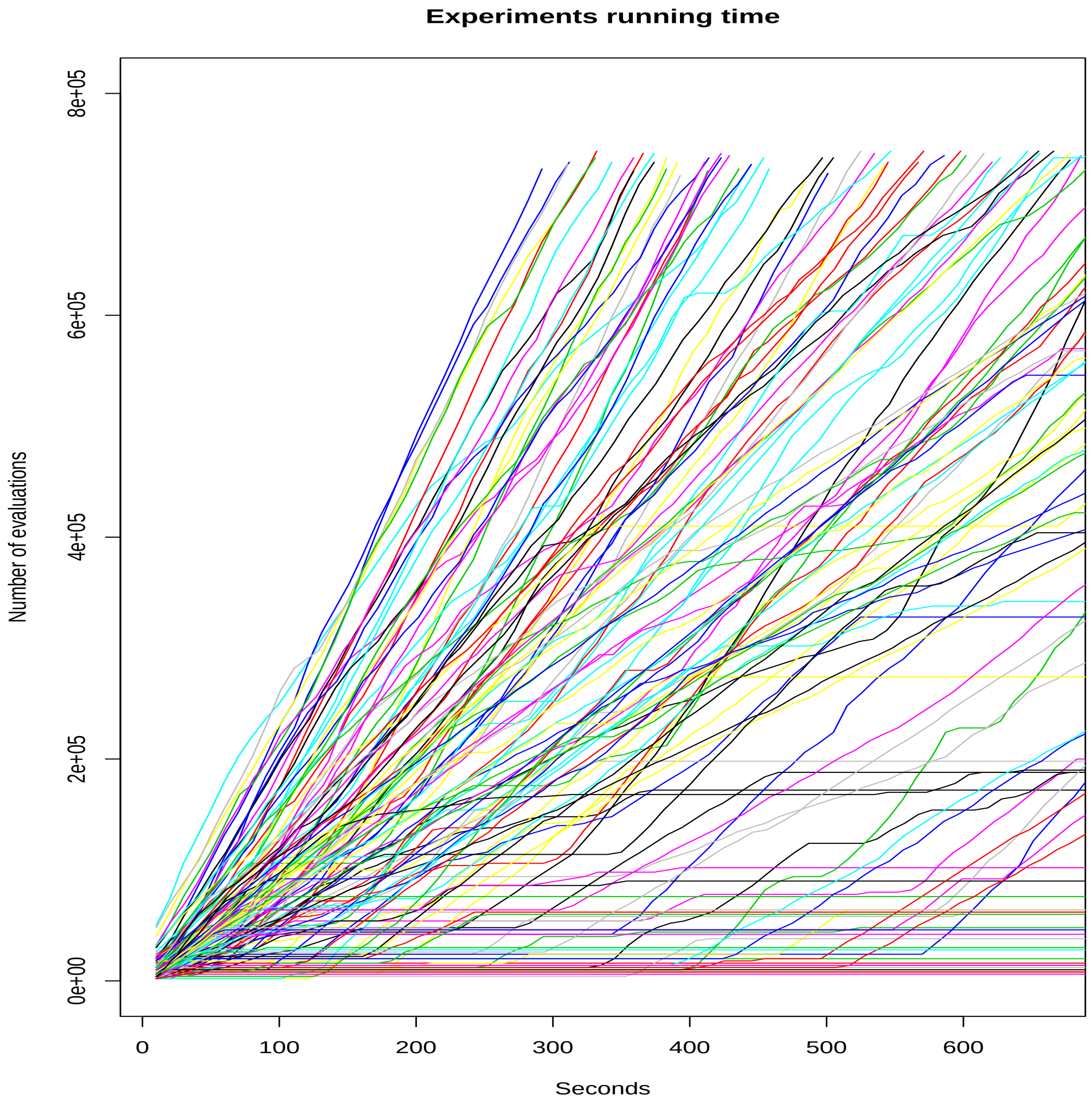

Fig. 6. Plot showing the number of individuals evaluated vs. time for a few dozens experiments; time is plotted up to 600 seconds only. As is seen, some experiments have a more or less constant evaluation rate (constant inclination), while other are more step-like with clients leaving and joining the experiment all the time.

The GA, being asynchronous, faces some problems that have not been tackled in this paper. What is the best approach to preserve diversity? To generate a new population in each client, and receive inmigrants as soon as possible, which are incorporated into the population? Or is it better to create new client populations based on existing populations? What is really the algorithmic contribution of new clients? These issues will be explored as future work. We will also try to measure the limits of this technology, and test the impact of servers of varying performance and workload on overall performance. Eventually, we will also try to perform a *sneaky* experiment, to check what kind of performance can be expected in that kind of setups.

Another venue of work will be to examine the algorithmic performance of AGAJAJ; even as new clients are added to an experiment, what's the improvement obtained from them? In order to check that, a controlled experiment using known computers will be used, adding them one at a time, so that the real impact on the genetic algorithm is evaluated. Once that is know, it would be interesting to experiment with adaptive client parameters, instead of the one-size-fits-all parameter settings used so far.

## Acknowledgements

This paper has been funded in part by the Spanish MICYT project NoHNES (Spanish Ministerio de Educación y Ciencia - TIN2007-68083) and the Junta de Andalucía P06-TIC-02025. We are also grateful to the editors of the

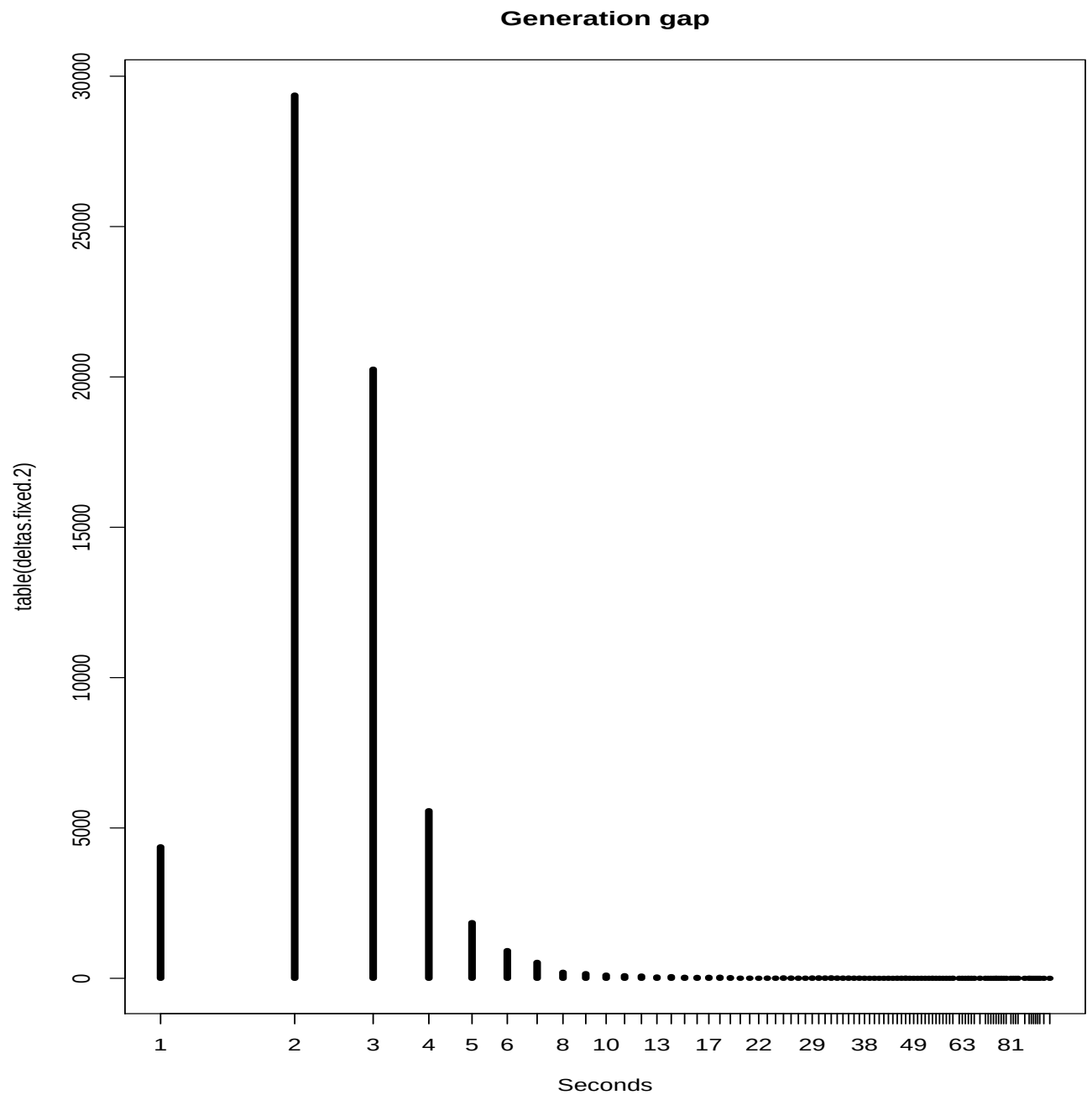

Fig. 4. Distribution of *gap* time, that is, time among two request from a particular client; it has been rounded to the closest number of seconds (since the webserver logs it has been extracted from have that resolution). Most clients take 3 second or less to process 20 generations. with a majority taking 2 or less. Gaps bigger than 10-15 seconds are probably non-significative, in fact, negative gaps and those bigger than 100 have been eliminated from the data set. For this set of experiments, the median is at 2. $x$ axis is logarithmic, to emphasize the fact that the distribution of client performance falls very fast, although a different center should be expected for a different problem.

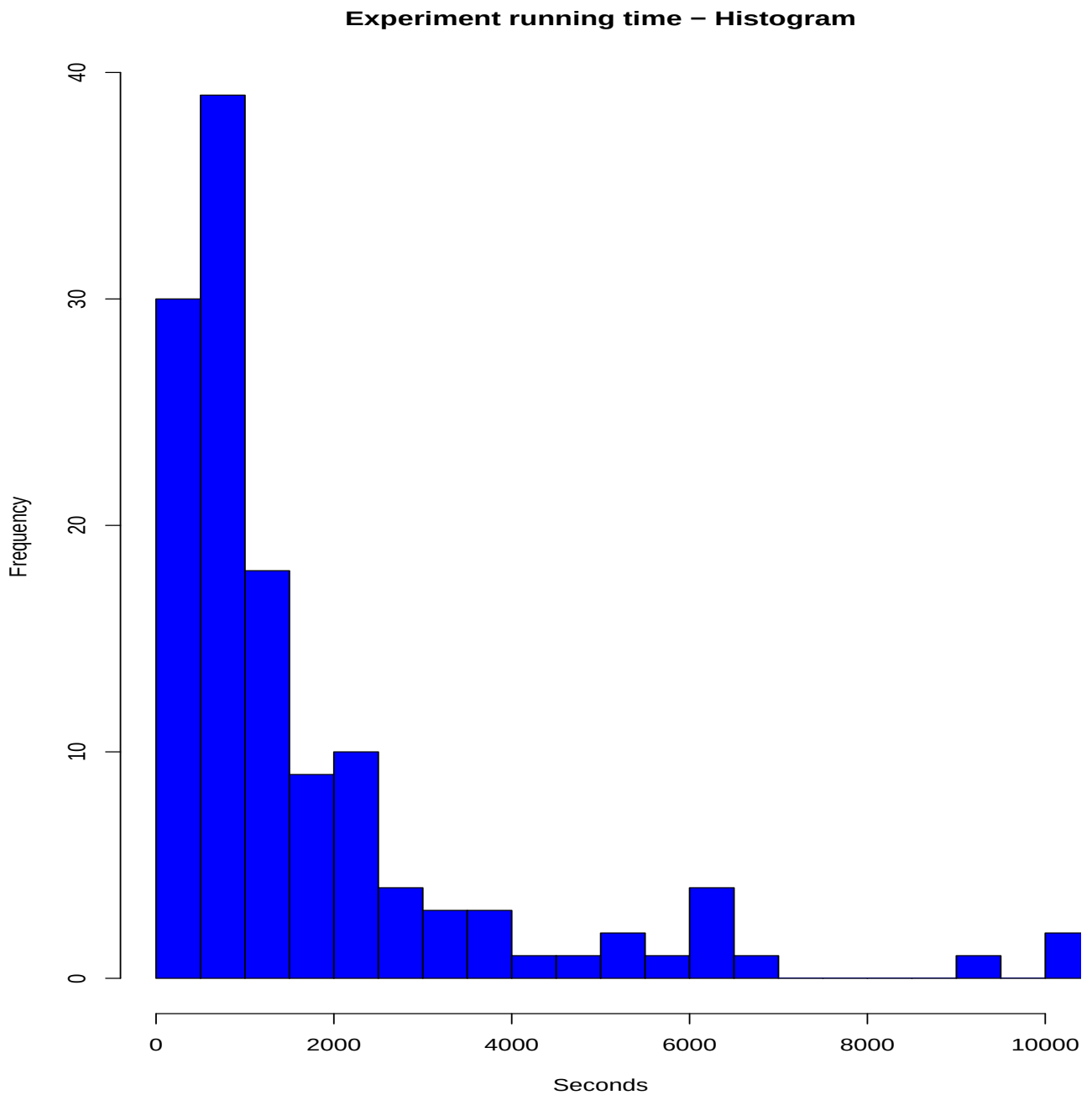

Fig. 5. Distribution of running times for a fixed amount of evaluations. Some outliers have been cut off; approximately 10% of runs took more than 10000 seconds. The mode is between 500 and 1000 seconds, and most runs end before 2000 seconds. Time starts to count from the moment the first client connects to the server.

Microsiervos.com blog for talking about our experiment, and to the readers of that article for participating in it. We are also grateful to the anonymous people who have known about the experiment via several possible ways[8] and participating in it.

[8]For instance, this article in my own blog `http://atalaya.blogalia.com/historias/53480`